\documentclass[final, journal, 10pt]{IEEEtran}

\usepackage[linesnumbered,lined,boxed,commentsnumbered]{algorithm2e}
\usepackage{algpseudocode}

\usepackage{amsmath,amsfonts,amssymb}     
\usepackage{graphicx}
\usepackage{csquotes}
\usepackage{bm}
\usepackage{subcaption}
\usepackage{xcolor}
\usepackage[english]{babel}
\newtheorem{theorem}{Theorem}
\newtheorem{proposition}{Proposition}
\newtheorem{lem}{Lemma}
\usepackage{floatrow}
\usepackage{multicol}
\usepackage{graphicx}
\usepackage{cite}

\title{Convexifying Sparse Interpolation with Infinitely Wide Neural Networks: An Atomic Norm Approach}
\author{Akshay Kumar\thanks{The authors are with the Department of Electrical and Computer Engineering, University of Minnesota, Minneapolis, MN, 55455 USA; e-mails: \{kumar511,jdhaupt@umn.edu\}. AK was supported in part by the 3M Science and Technology Doctoral Fellowship.}, and Jarvis Haupt, \IEEEmembership{Senior Member, IEEE}}
\begin{document}
	\maketitle
	\begin{abstract}
This work examines the problem of exact data interpolation via sparse (neuron count), infinitely wide, single hidden layer neural networks with leaky rectified linear unit activations. Using the atomic norm framework of [Chandrasekaran et al., 2012], we derive simple characterizations of the convex hulls of the corresponding atomic sets for this problem under several different constraints on the weights and biases of the network, thus obtaining equivalent convex formulations for these problems. A modest extension of our proposed framework to a binary classification problem is also presented. 
We explore the efficacy of the resulting formulations experimentally, and compare with networks trained via gradient descent.
\end{abstract}
\section{Introduction}
Deep learning has attracted significant attention in recent years, due in large part to its empirical success in numerous areas, including computer vision and natural language processing to name just a few. One of the salient characteristics shared among well-performing networks is that they are often heavily overparametrized, and despite having no explicit regularization, tend to have excellent generalization performance. 

It has been suggested (e.g., in \cite{bartlett_nips,norm_nati}) that for generalization, the norm(s) of the latent network parameters may be more important than the number of parameters. Motivated by this, there has been a growing interest lately on understanding \emph{infinitely wide} neural networks under norm constraints on their parameters; see \cite{cvx_nn, bach_cvx_nn, lin_srebro, ongie_min}.

In this work we consider the problem of fitting a given set of data by an infinitely wide neural network with a sparsity constraint to regularize the number of hidden neurons. We consider several different scenarios, characterized by different constraints on the weights and biases of the latent factors, and following the atomic norm framework \cite{atm_norm} we provide straightforward characterizations of
the convex hulls of the corresponding atomic sets.  Using these, we reformulate the original neural network interpolation problems as equivalent finite dimensional convex optimization problems.

Finally, we develop extensions of this framework to binary
classification tasks, and explore the efficacy of (all of) the resulting
convex formulations numerically through (admittedly limited, due to space constraints) experimental evaluations on low-dimensional problems.

\subsection{Problem Statement}
Suppose we have $N$ data points $\{\mathbf{x}_i,y_i\}_{i=1}^{N}$, where $\mathbf{x}_i\in\mathbb{R}^d$ and $y_i\in \mathbb{R}$. For a generic input $\mathbf{x}$, we let $f(\mathbf{x})$ denote the output of an infinitely wide neural network with a single hidden layer and normalized weights. Specifically, motivated by the formulation
in \cite{lin_srebro}, we define ${\mathcal W}\subset \mathbb{R}^d\times \mathbb{R}$, and write
\begin{equation}
    f(\mathbf{x}) = \int_{{\mathcal W}} \sigma(\mathbf{w}^T\mathbf{x}+b)d\mu(\mathbf{w},b) ,
\end{equation}
where $\mu$ is a signed measure on ${\mathcal W}$ and $\sigma(.)$ denotes the activation function. Here, our specific focus is on Leaky Rectified Linear Unit (Leaky ReLU) activations \cite{maas2013rectifier}; i.e.,
$$\sigma(x) = \left\{\begin{matrix}
x, \ \ \mbox{if} \ \  x\geq0,\\ 
\alpha x, \ \ \mbox{if} \ \  x<0
\end{matrix}\right.,$$
where $\alpha\neq1$. Note that $\alpha = 0$ corresponds to the ``traditional'' ReLU activation, and is subsumed in our more general model. 

Similar to \cite{bach_cvx_nn}, we consider the following\footnote{Recall that for any given measurable space $\left(\mathcal{X},\mathcal{M}\right)$ and signed measure $\nu$ defined on it, the total variation of $\nu$, denoted by $\|\nu\|_{TV}$, is defined as
$\|\nu\|_{TV} = \nu^++\nu^-,$
where $\nu^+$ and $\nu^-$ are unique positive measures such that $\nu = \nu^+ - \nu^-$ and $\nu^+\perp\nu^-$; see \cite{real_analysis}.  } optimization problem, whose aim is to interpolate the $N$ data points
\begin{align}
\label{l1_reg}
    &\min \|\mu\|_{TV} \\
    &\text{s.t.} \ \ y_i = \int_{{\mathcal W}} \sigma(\mathbf{w}^T\mathbf{x}_i+b)d\mu(\mathbf{w},b), \  \forall \ i=1,\dots,N.\nonumber
\end{align}
For Dirac measures, for which $\mu = \sum_{i=1}^m v_i\delta(\mathbf{w}_i,b
_i)$ for some $m\in\mathbb{N}$ and $\{\mathbf{w}_i,b_i\}_{i=1}^m$, $f(\mathbf{x})$ reduces to a finite width neural network,
$$f(\mathbf{x})=\sum_i^m v_i\sigma(\mathbf{w}_i^T\mathbf{x}+b_i),$$
and $\|\mu\|_{TV}$ reduces\footnote{For $p\geq 1$ the notation $\|\cdot\|_p$ denotes the $\ell_p$ norm of its vector argument.} to $\|\mathbf{v}\|_1$, where $\mathbf{v}=[v_1,\dots,v_m]$. As alluded above, our aim here is to reformulate variants of \eqref{l1_reg}  as equivalent convex optimization problems.

\section{Related works}
The use of sparsity inducing regularizers for selecting hidden neurons in infinitely wide single hidden layer neural networks was first studied in \cite{cvx_nn}; the generalization performance of related formulations was established in \cite{bach_cvx_nn}. 
For scalar inputs and ReLU activations, \cite{lin_srebro} establishes equivalence between minimum norm solutions and interpolating functions with minimum $\ell_1$ norms of their second derivatives. This direction is extended to multivariate inputs in \cite{ongie_min}. Although such function space perspectives provide useful insights, they essentially do not reduce the infinite dimensional nature of the original problem, instead transforming them into optimizations over specified function spaces.  Here, our formulation reduces these problems to finite dimensional convex optimizations.

Perhaps the work most closely related to ours is  \cite{pilanci_dual}, which considered optimizing finite but sufficiently wide neural networks with sparsity inducing regularization. Using semi-infinite duality, that work established  equivalence to convex optimization problems that are essentially regularized versions of the results we present here.  That said, our results hold for a broader class of activation functions, and we provide a more concise, atomic norm based proof technique.
\section{Main Results}
\subsection{Reformulation Under a Bounded Weights Condition}
We begin by establishing the atomic norm formulation of \eqref{l1_reg} using the framework outlined in  \cite{atm_norm}, for a special case where 
\begin{equation}\label{w_bound}
{\mathcal W}=\{\mathbf{w}\in\mathbb{R}^d, b\in\mathbb{R} : \|\mathbf{w}\|_2 \leq 1\}.
\end{equation}

Let $\mathbf{a}(\mathbf{w},b)\in\mathbb{R}^N$ denote the activation of a hidden neuron with parameters $(\mathbf{w},b)$ for all of the input data, so that
$$\mathbf{a}(\mathbf{w},b) = [
\sigma(\mathbf{w}^T\mathbf{x}_1+b)\dots 
\sigma(\mathbf{w}^T\mathbf{x}_N+b)
]^T,$$
and let $a_i(\mathbf{w},b)$ denote the i$th$ entry of $\mathbf{a}(\mathbf{w},b)$. We say the pair $(\mathbf{w},b)$ is \emph{active} at $\mathbf{x}_i$ if $\mathbf{w}^T\mathbf{x}_{i}+b>0$. Now, for $\lambda>0$ we define
$${\mathcal{T}}^+_\lambda = \{\mathbf{z}\in\mathbb{R}^N: \exists (\mathbf{w},b) \ s.t. \  \mathbf{z} = \mathbf{a}(\mathbf{w},b), \|\mathbf{w}\|_2\leq\lambda \},$$
and
$$\mathcal{T}^-_\lambda = \{\mathbf{z}\in\mathbb{R}^N: \exists (\mathbf{w},b)\ s.t.\  \mathbf{z} = -\mathbf{a}(\mathbf{w},b), \|\mathbf{w}\|_2\leq\lambda \}.$$
Setting $\lambda = 1$, we choose $\mathcal{T}^+_1\cup\mathcal{T}^-_1$ as the atomic set, reducing the problem in \eqref{l1_reg} (with the weight condition in \eqref{w_bound}) to that of minimizing the corresponding atomic norm:
\begin{equation}
    \label{atm_norm}
    \min t \ \mbox{s.t.} \  \mathbf{y} \in t \ \text{conv}(\mathcal{T}^+_1\cup\mathcal{T}^-_1).
\end{equation}

Because the atomic set $\mathcal{T}^+_1\cup\mathcal{T}^-_1$ has infinitely many atoms, finding an efficient description of the convex hull directly is difficult in general. Our main insight here is to divide the atomic set into a finite number of convex subsets, where each convex subset corresponds to a unique sign pattern of the elements of the atomic set. Then the convex hull of the atomic set is simply a convex combination of the convex subsets.

To formalize this, we first define
\begin{align*}
    \mbox{sign}(x) = \left\{\begin{matrix}
1,\ \mbox{if} \ x \geq 0 \\
-1,\ \mbox{if} \ x < 0
\end{matrix}\right.,
\end{align*}
and note that the number of distinct sign patterns present among all the elements in $\mathcal{T}^+_\lambda$ is finite. (Indeed, if we define $\mathcal{J} = \{\mbox{sign}(\mathbf{z})\in\mathbb{R}^N:\mathbf{z}\in\mathcal{T}^+_\lambda\}$ 

then $|\mathcal{J}|\leq 2^N$ holds trivially; further refinements of this bound are discussed later.)  Notice also that for every sign pattern present in $\mathcal{T}^+_\lambda$, the negation is present in $\mathcal{T}^-_\lambda$. 

Now, for each fixed element $\mathbf{s}\in\mathcal{J}$, let us denote by $\mathcal{A}_\mathbf{s}^\lambda$ (resp. $\mathcal{B}_\mathbf{s}^\lambda$), all elements in $\mathcal{T}^+_\lambda$ (resp. $\mathcal{T}^-_\lambda$) having the sign pattern $\mathbf{s}$ (resp. $-\mathbf{s}$); i.e.,
$$\mathcal{A}^\lambda_\mathbf{s} = \{\mathbf{z}\in\mathcal{T}^+_\lambda:\mbox{sign}(\mathbf{z}) = \mathbf{s} \},$$ and  $$\mathcal{B}^\lambda_\mathbf{s}=\{\mathbf{z}\in\mathcal{T}^-_\lambda:\mbox{sign}(\mathbf{z}) = -\mathbf{s}\}.$$
Further, for each $\mathbf{s}\in\mathcal{J}$ define $\mathbf{h}(\mathbf{s})\in\mathbb{R}^N$ with elements
\begin{align}
   \mathbf{h}_k(\mathbf{s}) = \left\{\begin{matrix}
1,\ \mbox{if}\ \mathbf{s}_k = 1 \\
\alpha,\ \mbox{if}\ \mathbf{s}_k = -1
\end{matrix}\right..  
\end{align}
The following result establishes the convexity of each of the sets $\mathcal{A}^\lambda_\mathbf{s}$ and $\mathcal{B}^\lambda_\mathbf{s}$, and describes a scaling property that will be useful for us in what follows.

\begin{lem}
\label{cvx_si}
The sets $\mathcal{A}^\lambda_\mathbf{s}$ and $\mathcal{B}^\lambda_\mathbf{s}$ are always convex for any fixed $\mathbf{s}\in\mathcal{J}.$ Further, if $\mathbf{z}\in\mathcal{A}^1_\mathbf{s}$ then $\lambda\mathbf{z}\in\mathcal{A}^\lambda_\mathbf{s}$ (and if $\mathcal{B}^1_\mathbf{s}$ then $\mathcal{B}^\lambda_\mathbf{s}$).
\end{lem}
\begin{IEEEproof}
Let $\mathbf{X} = [\mathbf{x}_1 \dots \mathbf{x}_N]\in\mathbb{R}^{d\times N}$ be a matrix whose columns are the data features. Any $\mathbf{z}\in\mathcal{A}^\lambda_\mathbf{s}$ will satisfy
 \begin{equation}
 \label{simp_A}
 \mathbf{z} = \mathbf{h}(\mathbf{s})\odot\left(\mathbf{X}^T\mathbf{w}+b\mathbf{1}\right) 
 \end{equation}
with $\mathbf{z}\odot\mathbf{s}\geq0$ and $\|\mathbf{w}\|_2\leq \lambda$, 
where $\odot$ represents the Hadamard (element-wise) product and $\mathbf{1}$ is a vector of ones. All the constraints are convex, thus $\mathcal{A}^\lambda_\mathbf{s}$ is a convex set (similarly for $\mathcal{B}^\lambda_\mathbf{s}$).

For the second claim, note that if $\mathbf{z}\in\mathcal{A}^1_\mathbf{s}$, then using \eqref{simp_A},
$ \lambda\mathbf{z} = \mathbf{h}(\mathbf{s})\odot\left(\mathbf{X}^T(\lambda\mathbf{w})+\lambda b\mathbf{1}\right)$ with $(\lambda\mathbf{z})\odot\mathbf{s}\geq0$ and $\|\mathbf{w}\|_2\leq 1$. 
Choosing, $\mathbf{\tilde{w}} = \lambda\mathbf{w}, \tilde{b} = \lambda b$ and again using (\ref{simp_A}), we get $\lambda\mathbf{z}\in\mathcal{A}^\lambda_\mathbf{s}.$ The result for $\mathbf{z}\in\mathcal{B}^1_\mathbf{s}$ follows similarly.
\end{IEEEproof}

We now establish that a union of the sets $\mathcal{A}^\lambda_\mathbf{s}$ and $\mathcal{B}^\lambda_\mathbf{s}$ provides a covering of the atomic set.

\begin{figure*}
    \centering
    \begin{subfigure}[b]{0.24\textwidth}
        \centering
        \includegraphics[width=\textwidth]{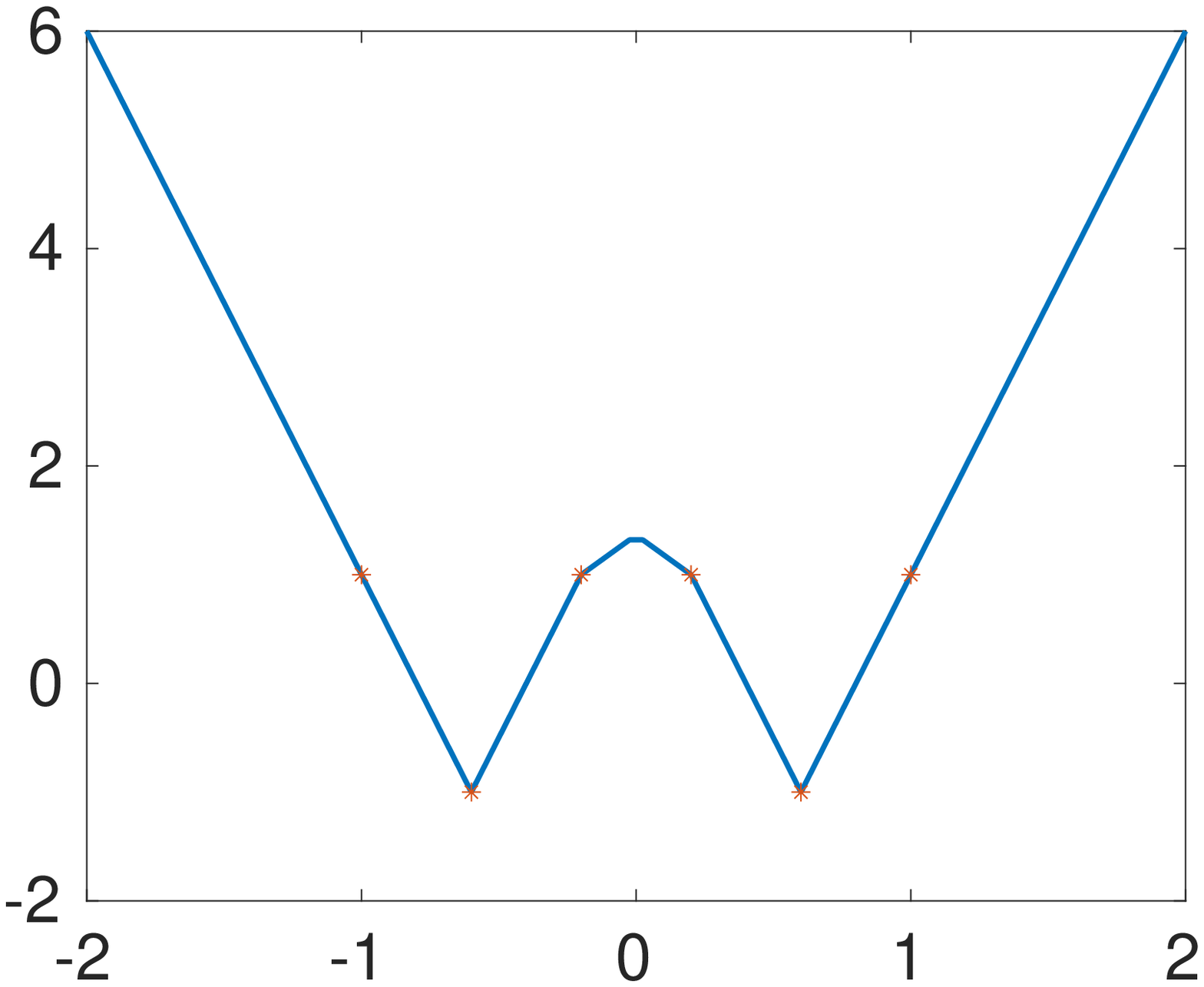}
        \caption[]{}%
        \label{cvx_sol}
    \end{subfigure}
    \begin{subfigure}[b]{0.24\textwidth}  
        \centering 
        \includegraphics[width=\textwidth]{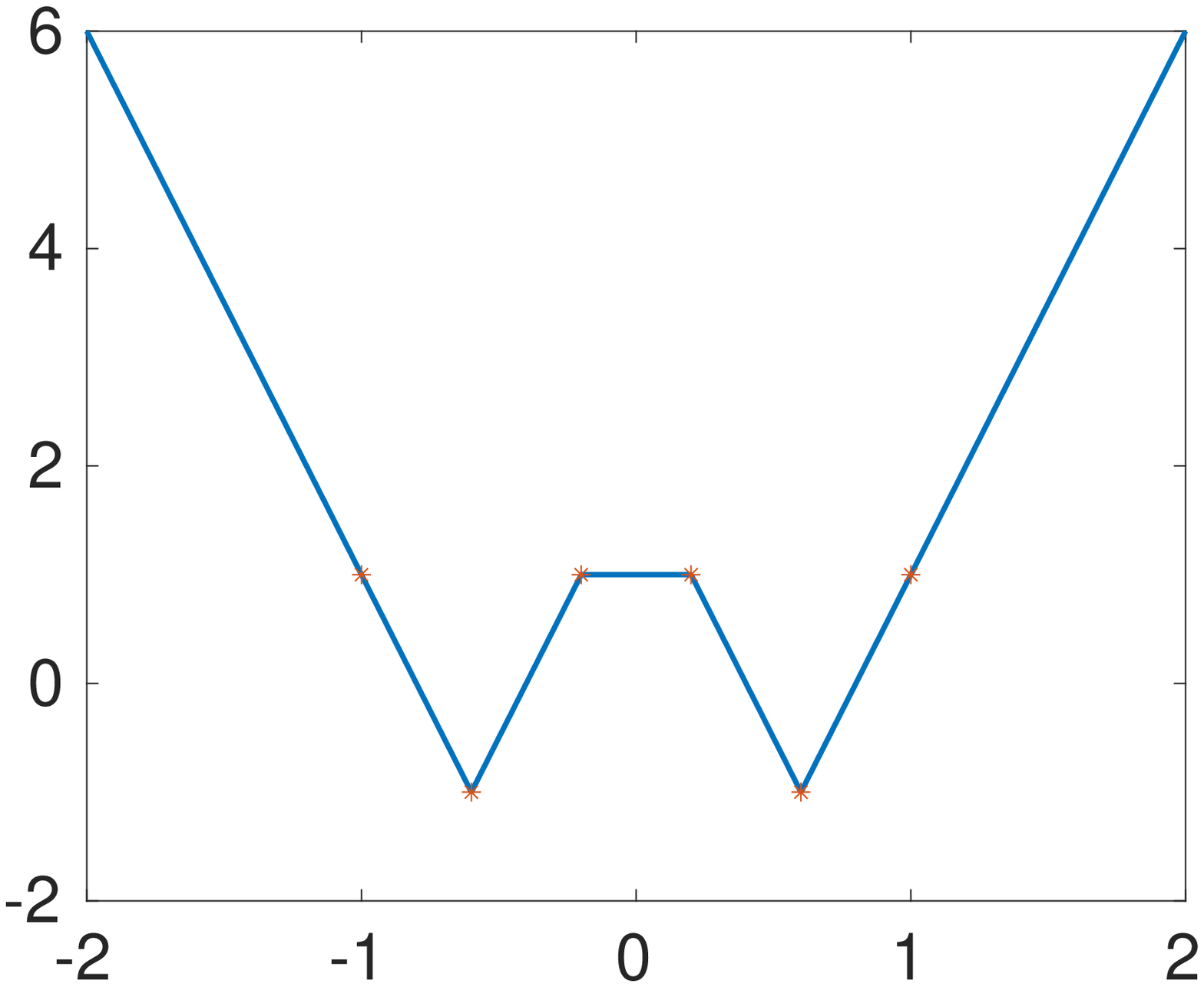}
        \caption[]{}%
        \label{lin_inter}
    \end{subfigure}
    \begin{subfigure}[b]{0.24\textwidth}   
        \centering 
        \includegraphics[width=\textwidth]{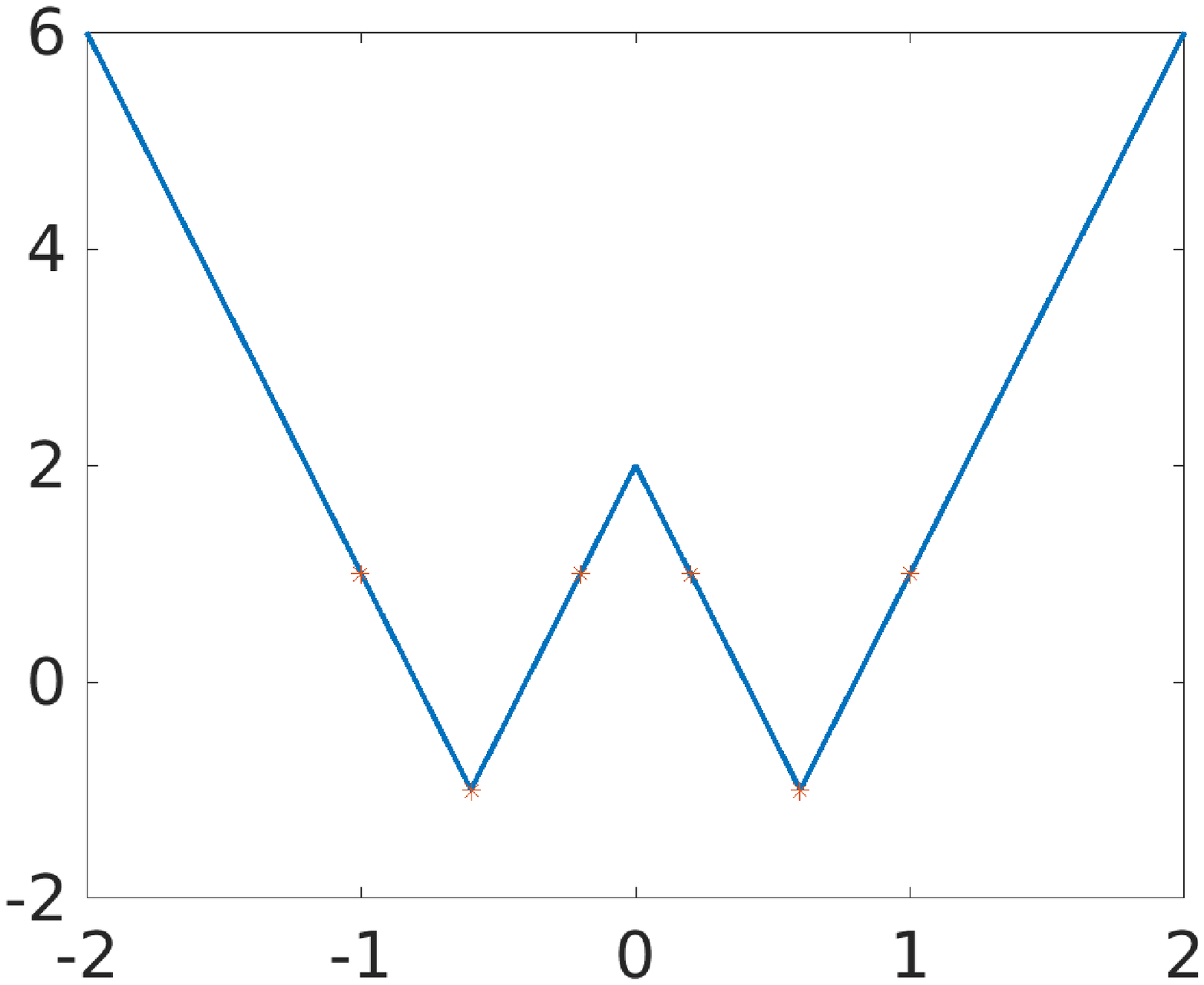}
        \caption[]{}%
        \label{sp_inter}
    \end{subfigure}
    \begin{subfigure}[b]{0.24\textwidth}   
        \centering 
        \includegraphics[width=\textwidth]{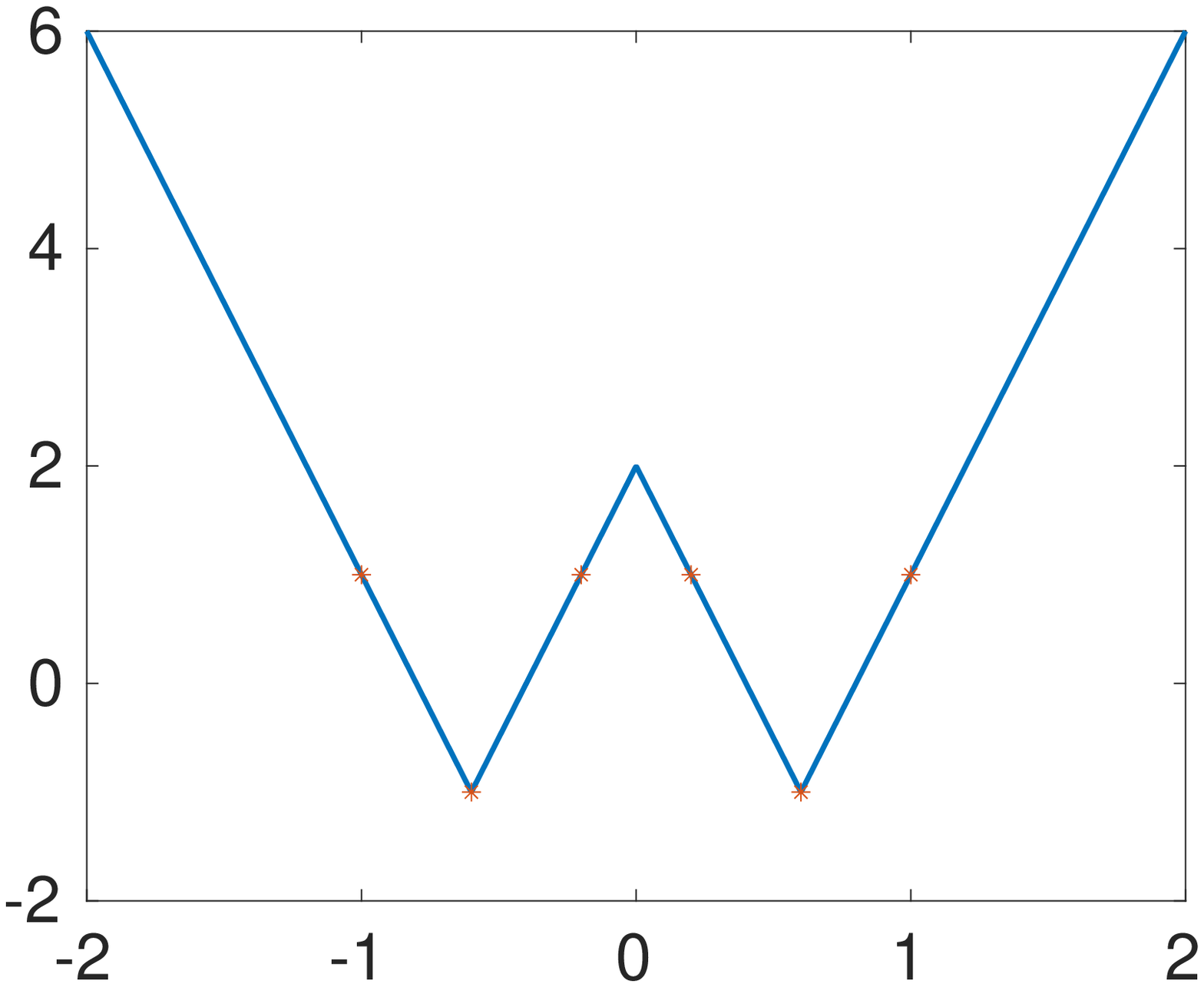}
        \caption[]{}%
        \label{cvx_bc_sol}
    \end{subfigure}\\
    \begin{subfigure}[b]{0.24\textwidth}
        \centering
        \includegraphics[width=\textwidth]{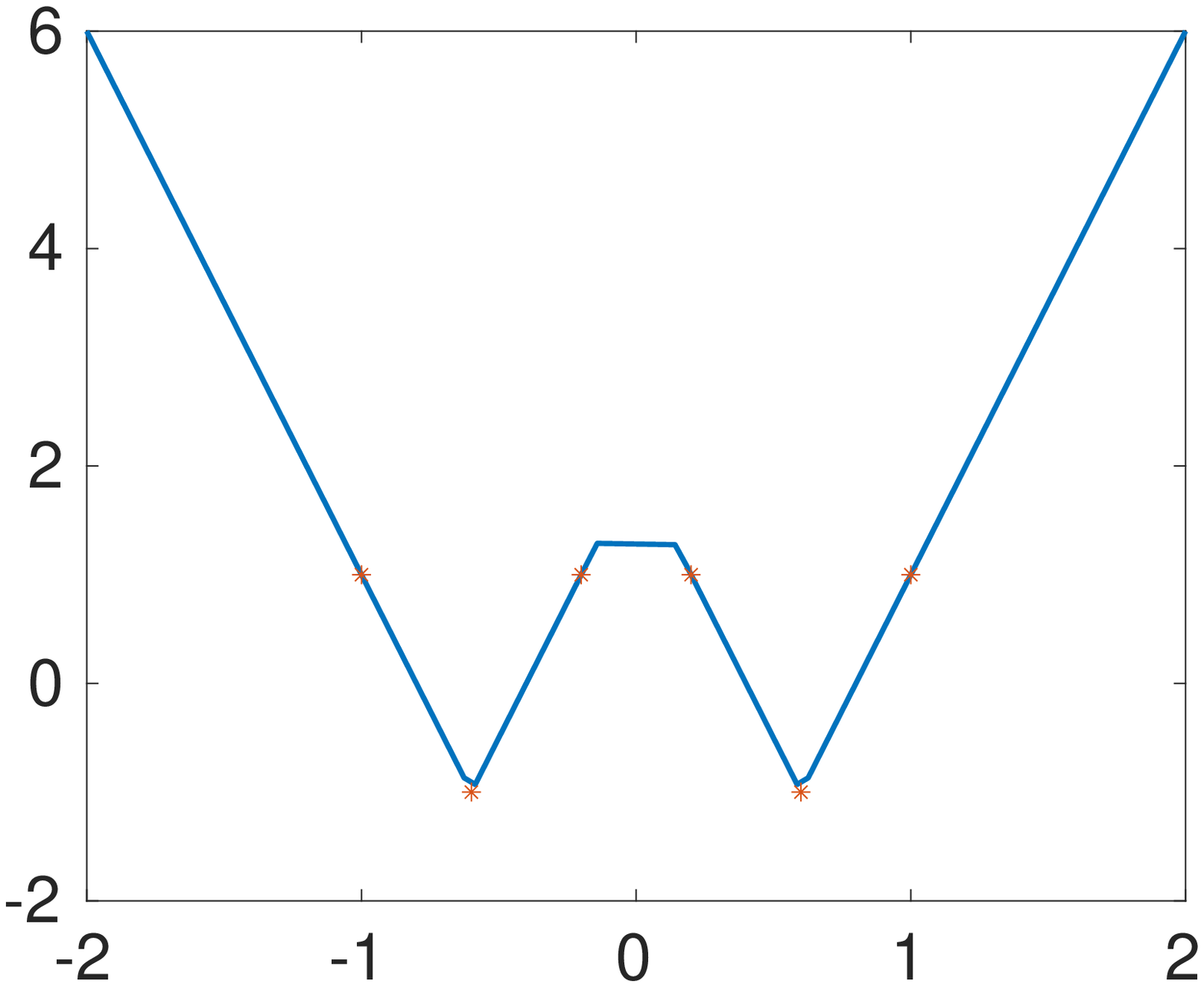}
        \caption[]{h = 500}%
        \label{dl_500}
    \end{subfigure}
    \begin{subfigure}[b]{0.24\textwidth}  
        \centering 
        \includegraphics[width=\textwidth]{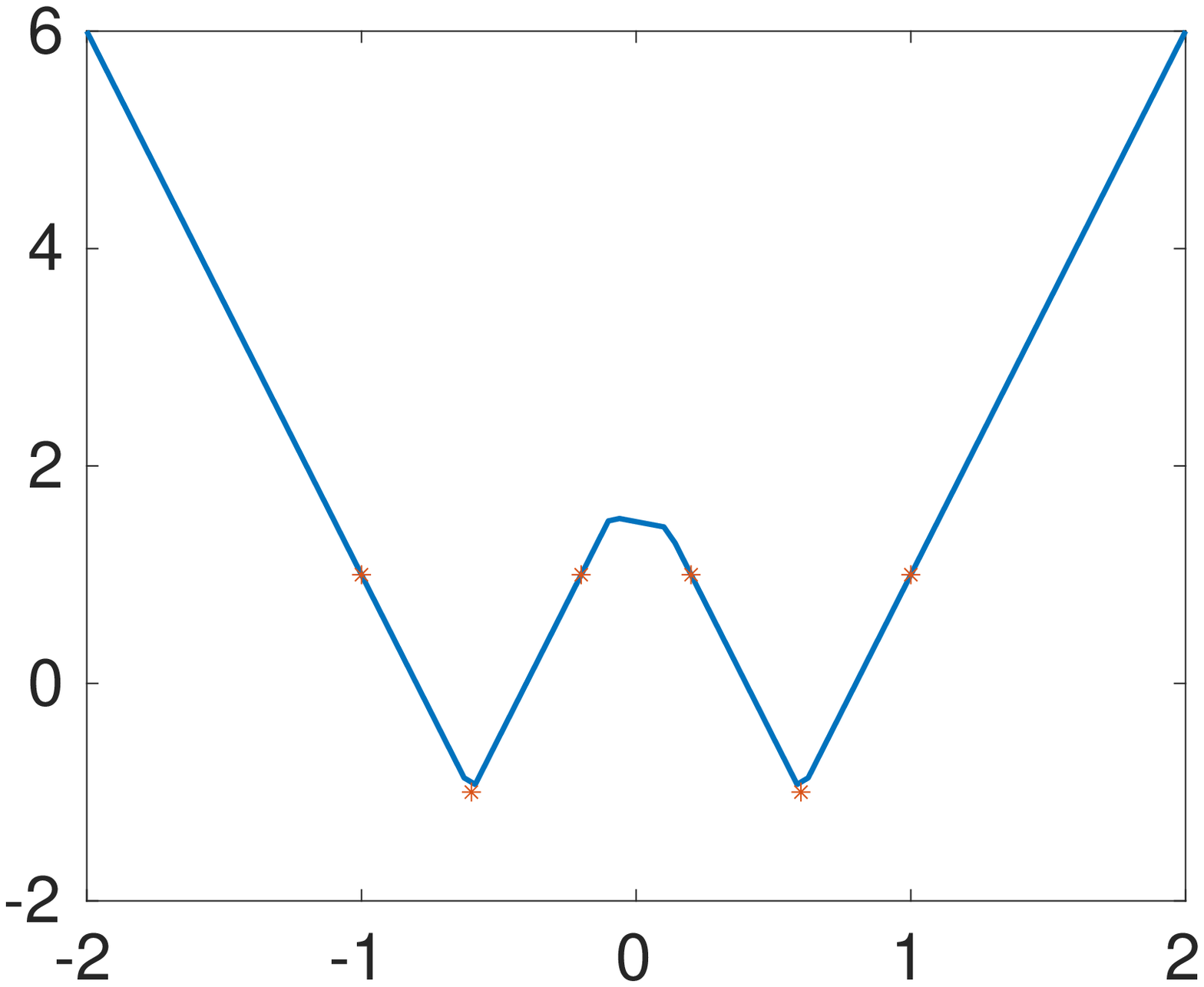}
        \caption[]{h = 1000}%
        \label{dl_1000}
    \end{subfigure}
    \begin{subfigure}[b]{0.24\textwidth}   
        \centering 
        \includegraphics[width=\textwidth]{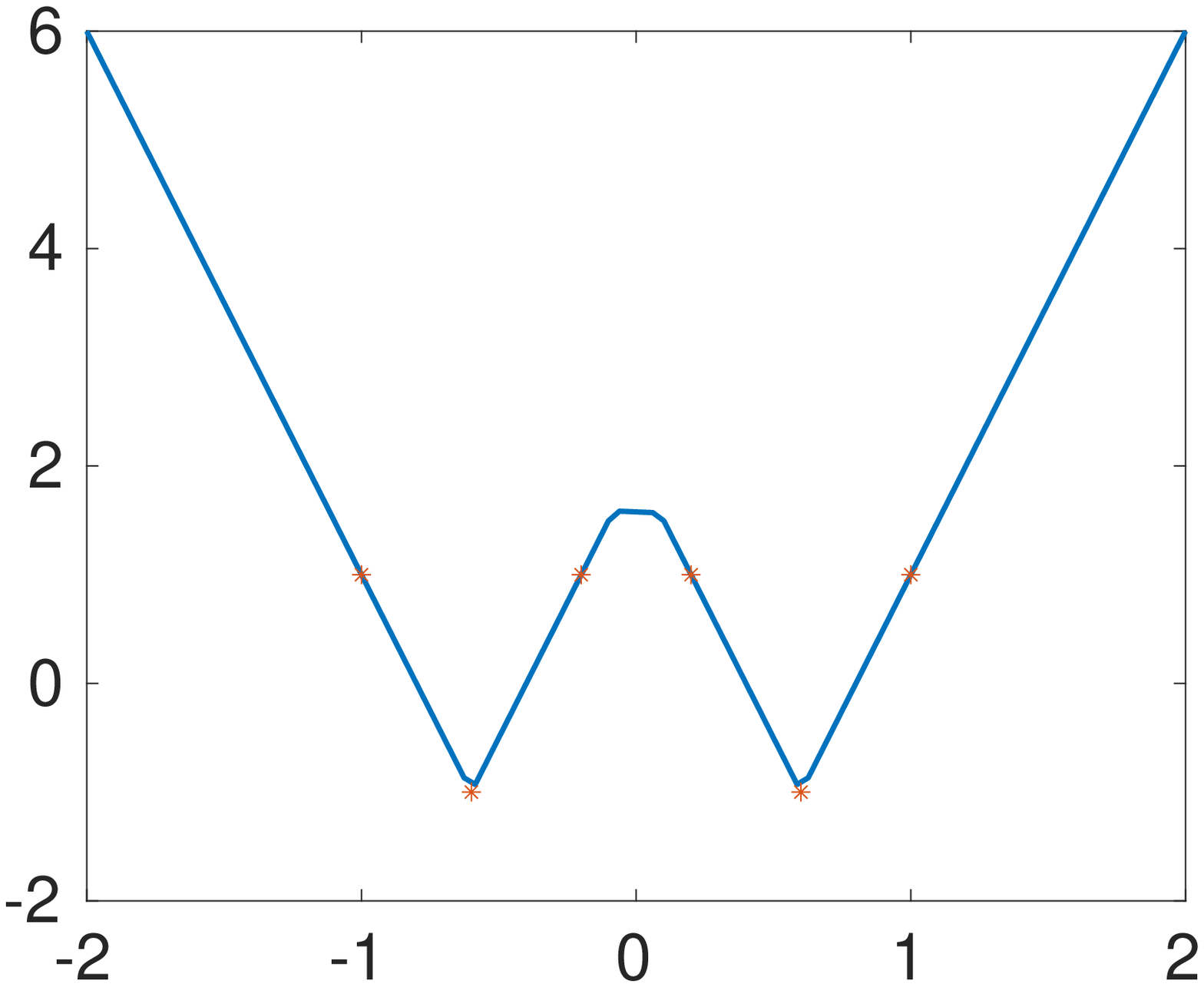}
        \caption[]{h = 2000}%
        \label{dl_2000}
    \end{subfigure}
    \begin{subfigure}[b]{0.24\textwidth}   
        \centering 
        \includegraphics[width=\textwidth]{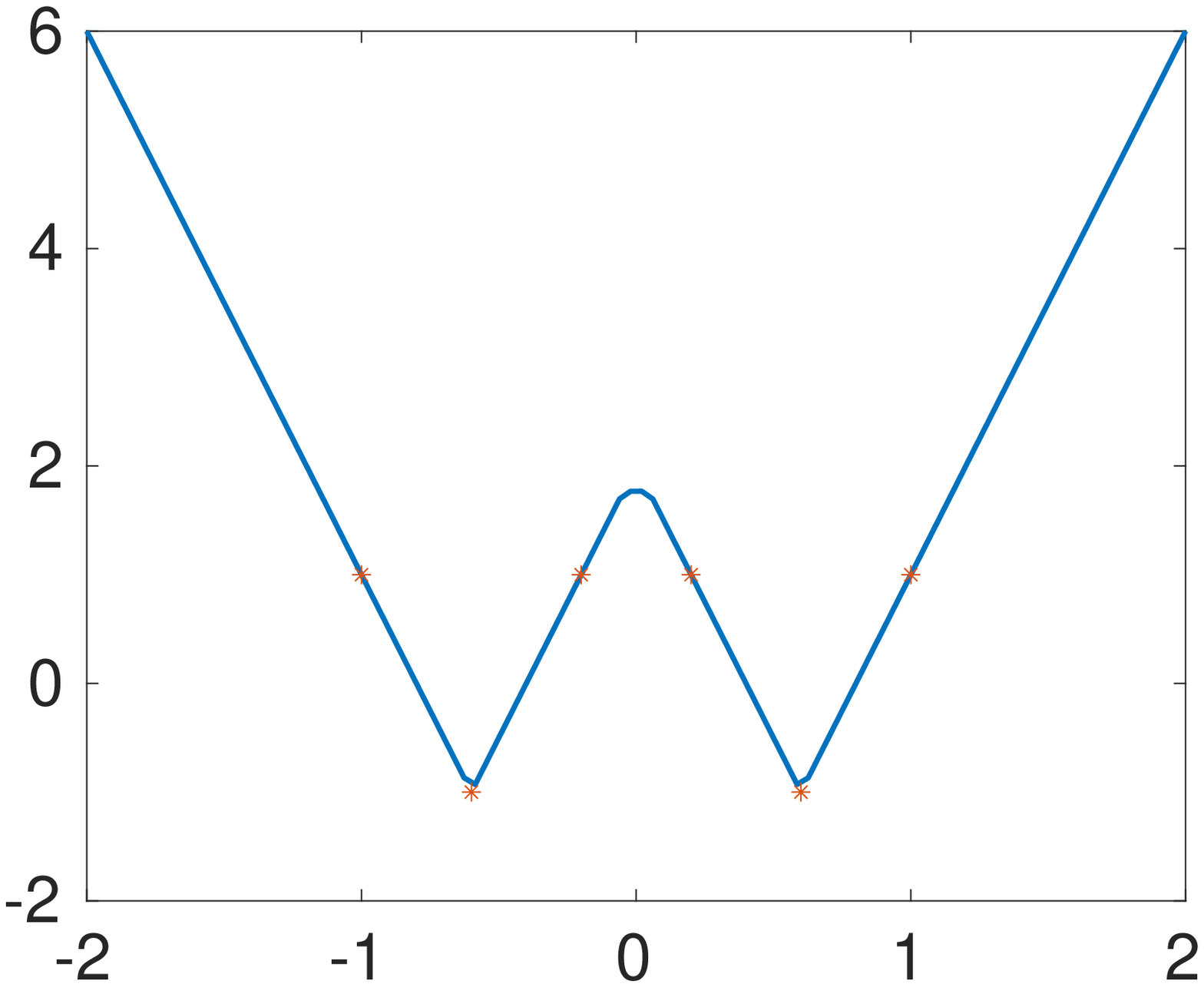}
        \caption[]{h = 5000}%
        \label{fig:dl_5000}
    \end{subfigure}
    \caption[]
    {\small a)-c) Different optimal solutions of (\ref{main_opt}). d) The solution of (\ref{mod_opt}). e)-f) The functions learned by neural networks with $h$ neurons trained by gradient descent, for varying values of $h$. }
    \label{fig:result}
\end{figure*}
\begin{lem}
\label{union}
We have $\mathcal{T}^+_\lambda = \bigcup_{i=1}^{|\mathcal{J}|}\mathcal{A}^\lambda_\mathbf{s_i}$, and $\mathcal{T}^-_\lambda = \bigcup_{i=1}^{|\mathcal{J}|}\mathcal{B}^\lambda_\mathbf{s_i}.$
\end{lem}
\begin{IEEEproof}
 First, it is clear that for any $\mathbf{s}_i\in\mathcal{J}$ we have $\mathcal{A}^\lambda_{\mathbf{s}_i}\subseteq\mathcal{T}^+_\lambda$, implying $\bigcup_{i=1}^{|\mathcal{J}|}\mathcal{A}^\lambda_{\mathbf{s}_i}\subseteq\mathcal{T}^+_\lambda$. 
On the other hand, for any $\mathbf{z}\in\mathcal{T}^+_\lambda$ let $\mathbf{s} = \mbox{sign}(\mathbf{z})$, then by definition of $\mathcal{J}$, $\mathbf{s}\in\mathcal{J}$. Hence, $\mathbf{z}\in\bigcup_{i=1}^{|\mathcal{J}|}\mathcal{A}^\lambda_{\mathbf{s}_i}$, implying $\mathcal{T}^+_\lambda\subseteq\bigcup_{i=1}^{|\mathcal{J}|}\mathcal{A}^\lambda_{\mathbf{s}_i}$, and proving the first part. The proof for $\mathcal{T}^-_\lambda$ is analogous.
\end{IEEEproof}

Now, using Lemmas \ref{cvx_si} and  \ref{union} above, as well as \cite[Theorem 3.3]{cvx_analysis}, we can obtain a finite representation of the convex hull of $\mathcal{T}^+_1\bigcup\mathcal{T}^-_1$, as stated in the following proposition.
\begin{proposition}
\label{cvx_hull}
Any $\mathbf{z}\in \mbox{conv}(\mathcal{T}^+_1\bigcup\mathcal{T}^-_1)$ can be written as $$\mathbf{z}=\sum_{i=1}^{|\mathcal{J}|} \lambda_i\mathbf{c}_i+\sum_{i=1}^{|\mathcal{J}|} \beta_i\mathbf{d}_i$$ for $\lambda_i,\beta_i\geq0$, where $\mathbf{c}_i\in\mathcal{A}^1_{\mathbf{s}_i}$, $\mathbf{d}_i\in\mathcal{B}^1_{\mathbf{s}_i}$, and $\sum_{i=1}^{|\mathcal{J}|} (\lambda_i + \beta_i) = 1$. 
\end{proposition}

Now, equipped with the results established above, we are in a position to state our first main result concerning a convex formulation of the optimization problem in \eqref{l1_reg} with bounded weight vectors.  
\begin{theorem}
\label{theorem_cvx}
The optimization problem in (\ref{l1_reg}) with ${\mathcal W}$ as in \eqref{w_bound} can equivalently be written as the following finite convex optimization problem
\begin{align}
\label{main_opt}
 &\min_{\mathbf{w}_i^+,\mathbf{w}_i^-,b_i^+,b_i^-} \  \sum_{i=1}^{|\mathcal{J}|} \left(\|\mathbf{w}_i^+\|_2+\|\mathbf{w}_i^-\|_2\right)\\
 &\mbox{s.t.}\ \mathbf{y} = \sum_{i=1}^{|\mathcal{J}|} \mathbf{h}(\mathbf{s}_i)\odot\left(\mathbf{X}^T(\mathbf{w}_i^+-\mathbf{w}_i^-)+(b_i^+-b_i^-)\mathbf{1}\right) \nonumber\\
 & \mathbf{s}_i\odot(\mathbf{X}^T\mathbf{w}_i^++b_i^+\mathbf{1})\geq0 \ \forall \mathbf{s}_i\in\mathcal{J}\nonumber \\
 & \mathbf{s}_i\odot(\mathbf{X}^T\mathbf{w}_i^-+b_i^-\mathbf{1})\leq0 \ \forall \mathbf{s}_i\in\mathcal{J},\nonumber
\end{align}
where $\mathbf{y}=[y_1 \dots y_N]\in\mathbb{R}^N$ is the vector of data point labels.
\end{theorem}
\begin{IEEEproof}
Using Proposition \ref{cvx_hull}, \eqref{atm_norm} can be written as
\begin{align*}
 &\min_{t, \{\hat{\lambda}_i,\hat{\beta}_i\}_{i=1}^{|\mathcal{J}|}} \ t\\
 & \mbox{s.t.}\ \mathbf{y} = \sum_{i=1}^{|\mathcal{J}|} t\hat{\lambda}_i\mathbf{c}_i+\sum_{i=1}^{\mathcal{J}} t\hat{\beta}_i\mathbf{d}_i, \\
 & \mathbf{c}_i\in\mathcal{A}^1_{\mathbf{s}_i}, \mathbf{d}_i\in\mathcal{B}^1_{\mathbf{s}_i},\\
 &\sum_{i=1}^{|\mathcal{J}|} (\hat{\lambda}_i+\hat{\beta}_i)=1,\ \ \hat{\lambda}_i,\hat{\beta}_i\geq0.
\end{align*}

Choosing, $\lambda_i = t\hat{\lambda}_i, \beta_i = t\hat{\beta}_i$ and eliminating $t$ this becomes
\begin{align*}
 &\min_{\{{\lambda}_i,{\beta}_i\}_{i=1}^{|\mathcal{J}|}} \ \sum_{i=1}^{|\mathcal{J}|} ({\lambda}_i+{\beta}_i)\\
 & \mbox{s.t.}\ \mathbf{y} = \sum_{i=1}^{|\mathcal{J}|} {\lambda}_i\mathbf{c}_i+\sum_{i=1}^{\mathcal{J}} {\beta}_i\mathbf{d}_i, \\
 & \mathbf{c}_i\in\mathcal{A}^1_{\mathbf{s}_i}, \mathbf{d}_i\in\mathcal{B}^1_{\mathbf{s}_i},\\
 &{\lambda}_i,{\beta}_i\geq0.
\end{align*}
Now, if $\mathbf{c}_i\in\mathcal{A}^1_{\mathbf{s}_i}$ and $\mathbf{d}_i\in\mathcal{B}^1_{\mathbf{s}_i}$, then by Lemma \ref{cvx_si} we have $\lambda_i\mathbf{c}_i\in\mathcal{A}^{\lambda_i}_{\mathbf{s}_i}$ and $\beta_i\mathbf{d}_i\in\mathcal{B}^{\beta_i}_{\mathbf{s}_i}$. Thus, using \eqref{simp_A}, we see that the optimization is equivalent to
\begin{align*}
 &\min_{\{{\lambda}_i,{\beta}_i\}_{i=1}^{|\mathcal{J}|}} \ \sum_{i=1}^{|\mathcal{J}|} ({\lambda}_i+{\beta}_i)\\
 & \mbox{s.t.}\ \mathbf{y} = \sum_{i=1}^{|\mathcal{J}|} \mathbf{h}(\mathbf{s}_i)\odot\left(\mathbf{X}^T(\mathbf{w}_i^+-\mathbf{w}_i^-)+(b^+-b^-)\mathbf{1}\right), \\
 & \mathbf{s}_i\odot(\mathbf{X}^T\mathbf{w}_i^++b^+\mathbf{1})\geq0, \|\mathbf{w}^+_i\|_2\leq\lambda_i,\\ &\mathbf{s}_i\odot(\mathbf{X}^T\mathbf{w}_i^-+b^-\mathbf{1})\leq0,\|\mathbf{w}^-_i\|_2\leq\beta_i,\\
 &{\lambda}_i,{\beta}_i\geq0, \forall i.
\end{align*}
Finally, eliminating $\lambda_i \ \text{and} \  \beta_i$, we arrive at the result.
\end{IEEEproof}

Note that the computational complexity of the (convex) optimization problem in Theorem~\ref{theorem_cvx} depends on the cardinality of $\mathcal{J}$. Recall that the set $\mathcal{J}$ contains the distinct sign patterns achievable by partitioning of the data points via hyperplanes. A trivial upper bound is $2^N$, but this can be refined  to $|\mathcal{J}| \leq 2\sum_{i=1}^d \binom{N}{i}$ (using \cite[Theorem 1]{cover}); in either case, the bound is exponential (in the number of data points in the former case, or in the input dimension in the latter).  Thus, for ``typical'' learning problems where the number of training data points $N$ is larger than the input dimension $d$, the formulation outlined above is intractable for even moderate choices of $d$. 
\subsection{Formulation Under Constrained Weights and Biases}
We next consider a variant of the problem in \eqref{l1_reg} where 
\begin{equation}\label{w_b_bound}
    {\mathcal W} = \{\mathbf{w}\in\mathbb{R}^d, b\in\mathbb{R}: \|[\mathbf{w}, b]\|_2 \leq 1\}.
\end{equation}
This formulation differs from the above in that the weight vector and the bias are jointly constrained inside an $\ell_2$ ball. 

To derive the corresponding convex optimization, the atomic set can be appropriately modified to incorporate the joint constraint and then the same technique outlined above can be employed. The proof is very similar so we omit it and state the result directly.
\begin{theorem}
The optimization problem in \eqref{l1_reg} with ${\mathcal W}$ as in \eqref{w_b_bound} can equivalently be written as following finite convex optimization problem,
\begin{align}
\label{mod_opt}
 &\min_{\mathbf{w}_i^+,\mathbf{w}_i^-,b_i^+,b_i^-} \  \sum_{i=1}^{|\mathcal{J}|} \left(\|[\mathbf{w}_i^+, \ b_i^+]\|_2+\|[\mathbf{w}_i^-,\ b_i^-]\|_2\right)\\
 & \mbox{s.t.}\ \mathbf{y} = \sum_{i=1}^{|\mathcal{J}|} \mathbf{h}(\mathbf{s}_i)\odot\left(\mathbf{X}^T(\mathbf{w}_i^+-\mathbf{w}_i^-)+(b_i^+-b_i^-)\mathbf{1}\right) \nonumber\\
 & \mathbf{s}_i\odot(\mathbf{X}^T\mathbf{w}_i^++b_i^+\mathbf{1})\geq0 \ \forall \mathbf{s}_i\in\mathcal{J}\nonumber \\ &\mathbf{s}_i\odot(\mathbf{X}^T\mathbf{w}_i^-+b_i^-\mathbf{1})\leq0 \ \forall \mathbf{s}_i\in\mathcal{J}.\nonumber
\end{align} 
\end{theorem}
Note that the optimization problems in ($\ref{main_opt}$) and (\ref{mod_opt}) are similar, but can have different solution sets, as will be shown in the experiments section.
\begin{figure}
  \begin{subfigure}{0.45\columnwidth}
  \includegraphics[width=\textwidth]{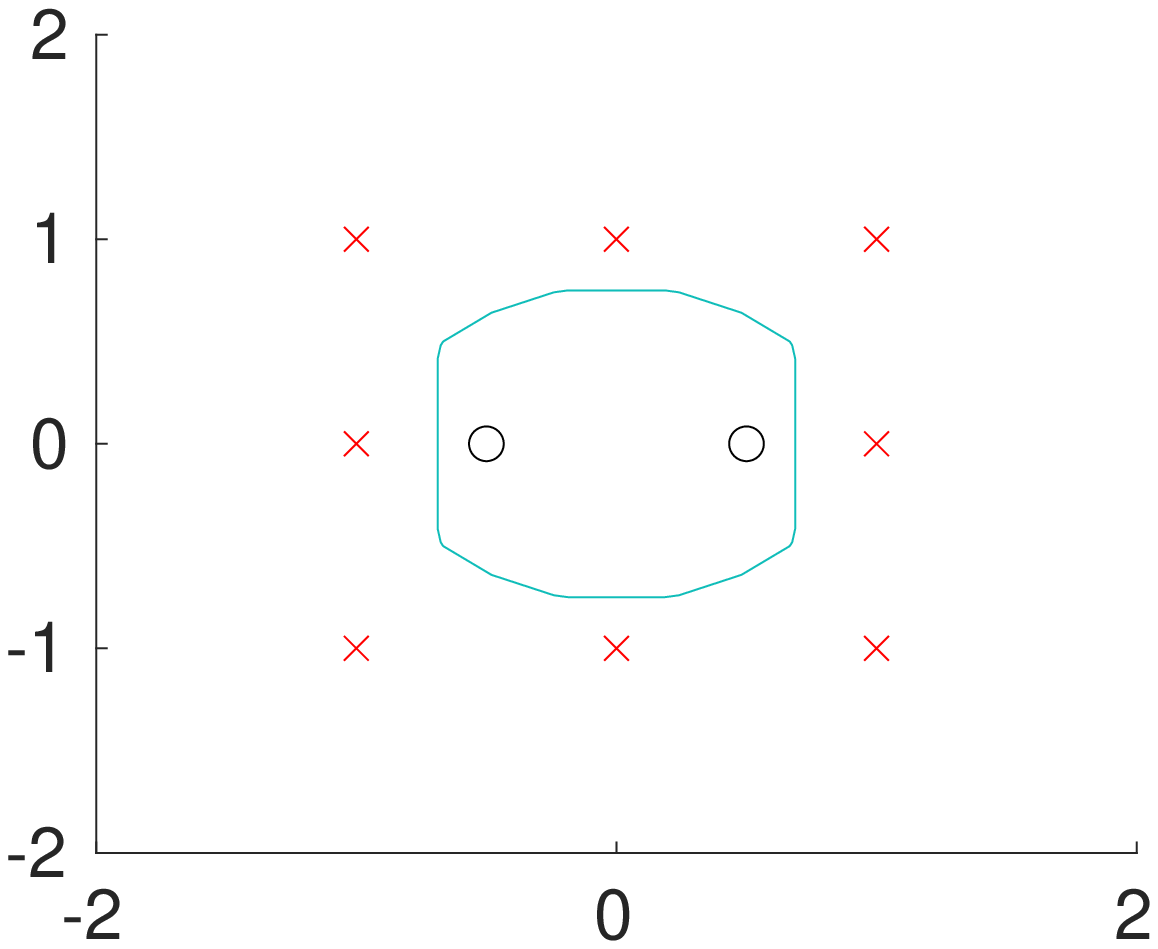}
   \caption{}
  \end{subfigure}
  \hfill
  \begin{subfigure}{0.45\columnwidth}
  \includegraphics[width=\textwidth]{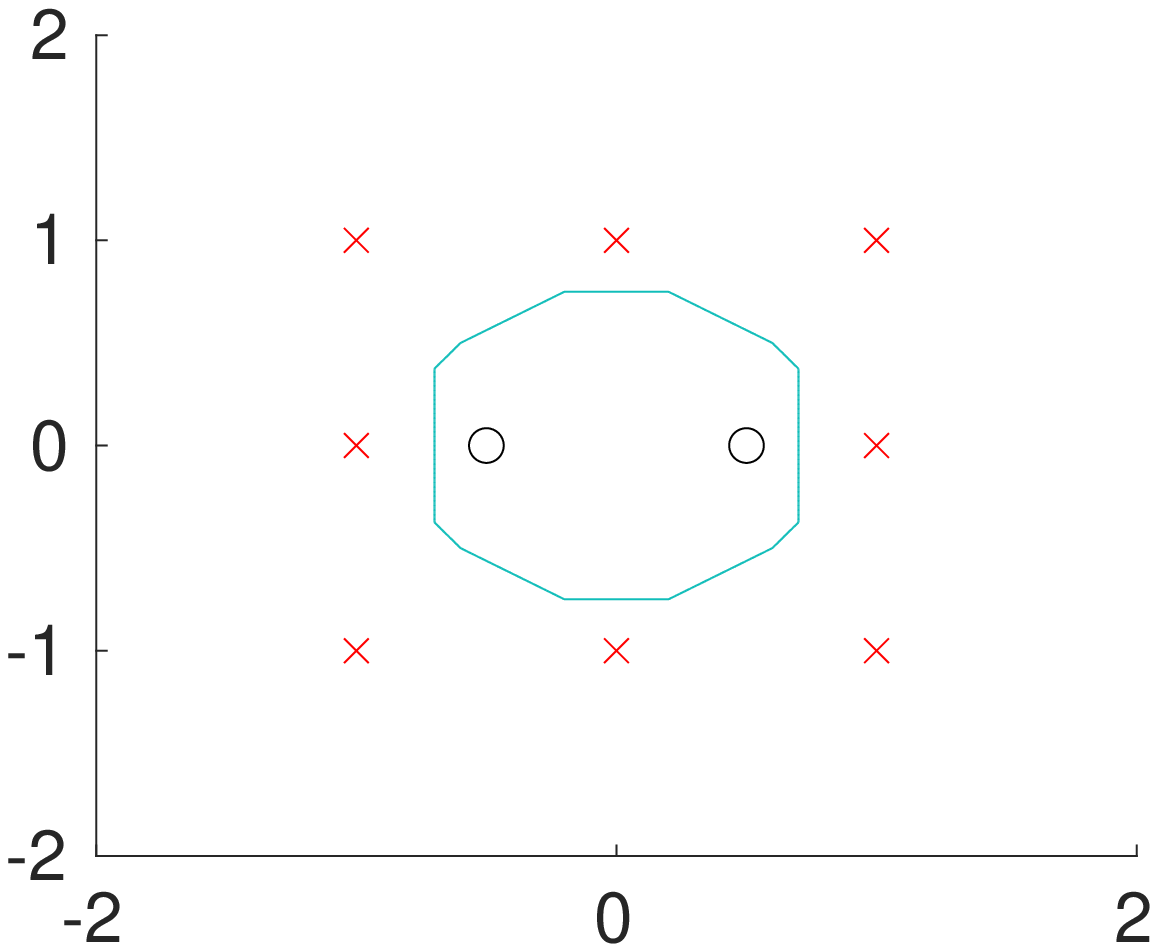}
   \caption{} 
  \end{subfigure} 
  \begin{subfigure}{0.45\columnwidth} 
  \includegraphics[width=\textwidth]{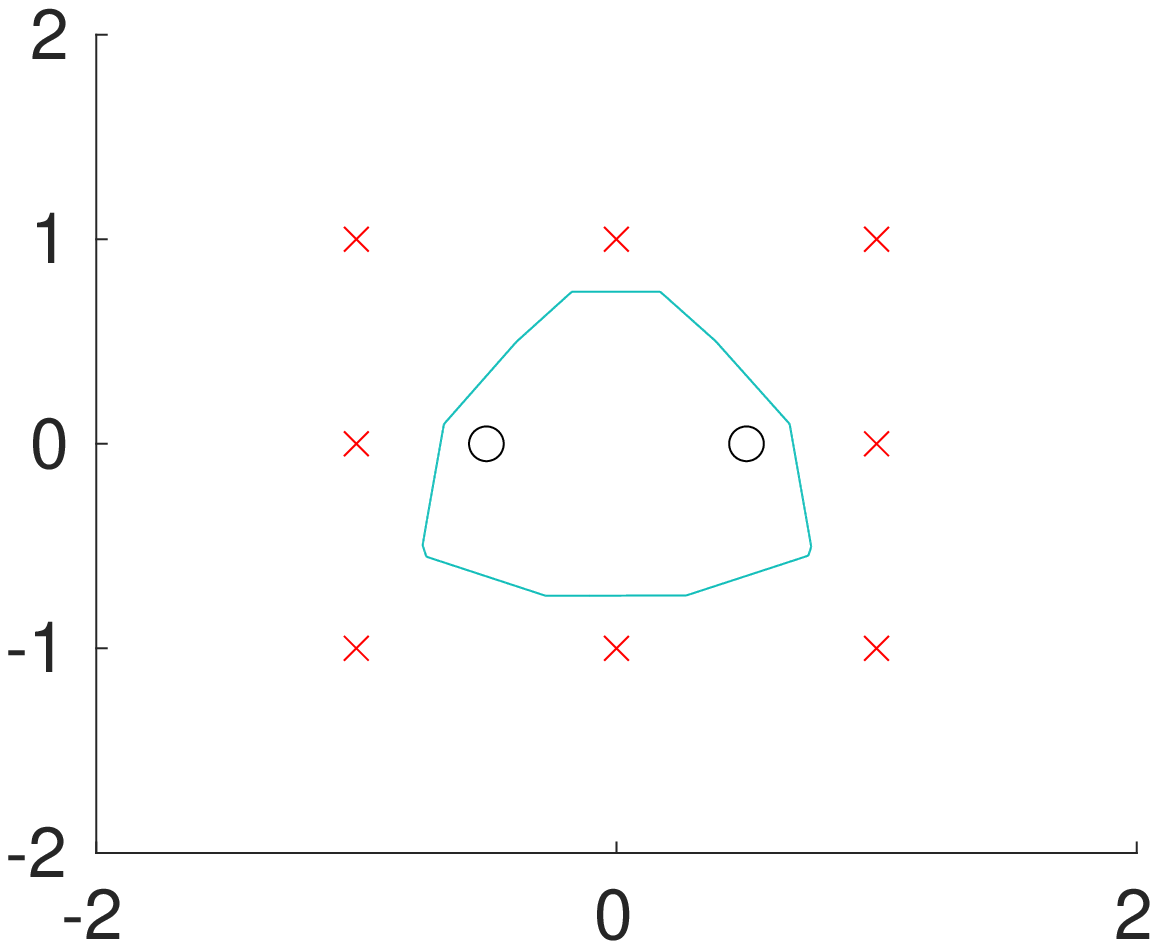} 
  \caption{h=1000} 
  \end{subfigure}  
  \hfill 
  \begin{subfigure}{0.45\columnwidth} 
  \includegraphics[width=\textwidth]{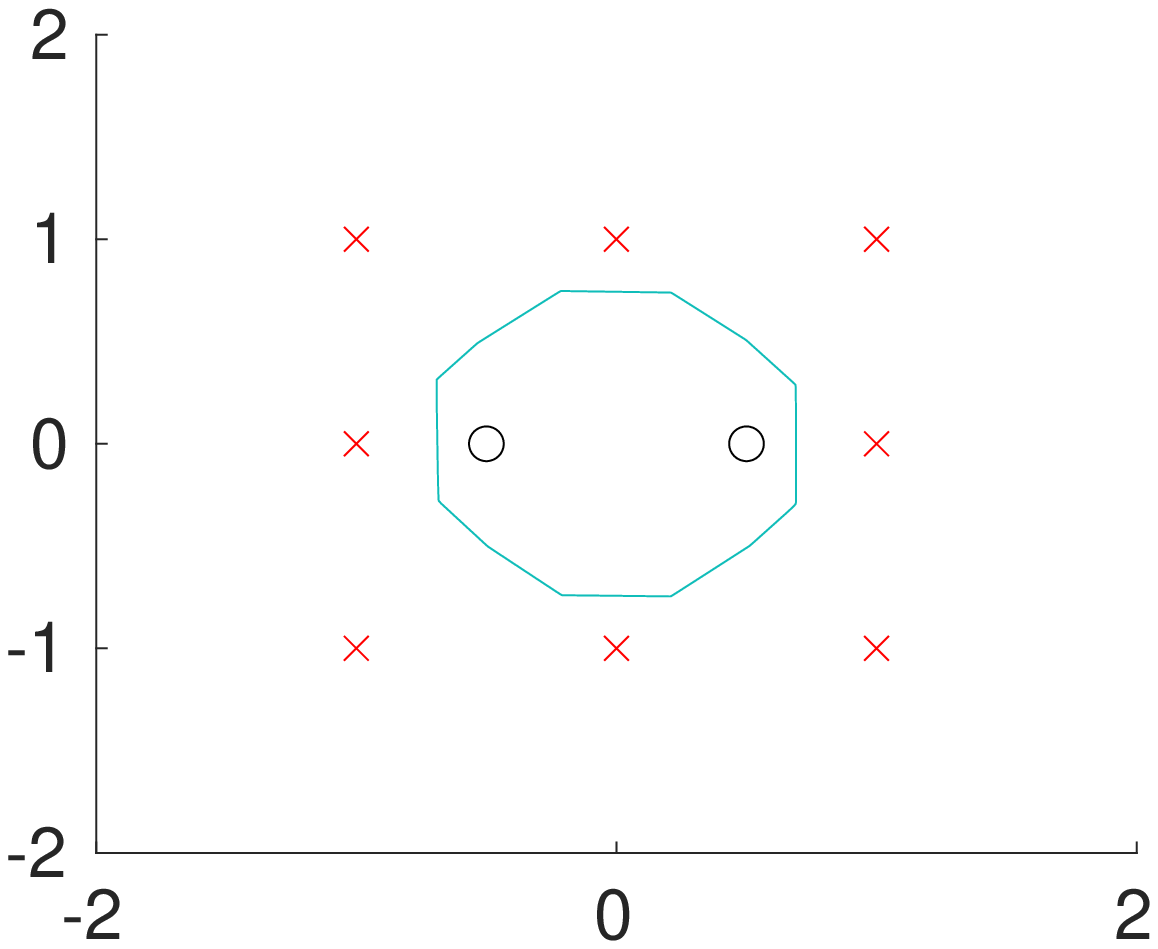} 
  \caption{h=5000} 
  \end{subfigure}
  \caption{a),b) Classifiers identified via convex formulations. c),d) Neural network solution with $h$ neurons trained by gradient descent.}
  \label{fig:result_bc}
  \end{figure}
\section{An Extension to Binary Classification}
An important variant of the above is the problem of binary classification, for which the corresponding minimum norm problem can be written as
\begin{align}
    \label{l1_bc}
    &\min \|\mu\|_{TV} \\
    &\text{s.t.} \ \ \int_{{\mathcal W}}y_i\sigma(\mathbf{w}^T\mathbf{x_i}+b)d\mu(\mathbf{w},b)\geq 1,  \forall i=1,\dots,N.\nonumber
\end{align}
To address this problem, we change the activation $\mathbf{a}(\mathbf{w},b)$ slightly and define it as
$$\mathbf{a}(\mathbf{w},b) = \begin{bmatrix}
y_1\sigma(\mathbf{w}^T\mathbf{x}_1+b) \dots 
y_N\sigma(\mathbf{w}^T\mathbf{x}_N+b)
\end{bmatrix}^T.$$
Building the atomic set using this modified activation and following the similar technique as for proof of Theorem $\ref{theorem_cvx}$, we can obtain the following result. 
\begin{theorem}
The optimization problem in (\ref{l1_bc}) can equivalently be written as following finite convex optimization problem,
\begin{align}
\label{bc_opt}
 &\min_{\mathbf{w}_i^+,\mathbf{w}_i^-,b_i^+,b_i^-} \ \sum_{i=1}^{|\mathcal{J}|} (\|\mathbf{w}_i^+\|_2+\|\mathbf{w}_i^-\|_2)\\
 & \mbox{s.t.}\ \sum_{i=1}^{|\mathcal{J}|} \mathbf{h(s_i)}\odot\left(\mathbf{y}\odot\left(\mathbf{X}^T(\mathbf{w}_i^+-\mathbf{w}_i^-)+(b_i^+-b_i^-)\mathbf{1}\right)\right)\geq 1 \nonumber\\
 & \mathbf{s}_i\odot(\mathbf{y}\odot\left(\mathbf{X}^T\mathbf{w}_i^++b_i^+\mathbf{1}\right))\geq0 \ \forall \  \mathbf{s_i}\in\mathcal{J}\nonumber\\ &\mathbf{s}_i\odot(\mathbf{y}\odot\left(\mathbf{X}^T\mathbf{w}_i^-+b_i^-\mathbf{1}\right))\leq0 \ \forall \  \mathbf{s_i}\in\mathcal{J}.\nonumber
\end{align} 
\end{theorem}
In the following section we evaluate each of these formulations numerically.

\section{Experiments}
We first examine the performance of the convex formulations for the interpolation problems, in settings where the input data are one dimensional ($d=1$).  Throughout, we use the standard ReLU activation, and solve the optimization problems in MATLAB using the CVX software \cite{cvx}. 

Due to space constraints, we consider here only an illustrative example characterized by $\mathbf{X} = [-1,-0.6,-0.2,0.2,0.6,1],$ and $\mathbf{y} = [1,-1,1,1,-1,1]$. In Figure \ref{fig:result}(a)-(c) we plot multiple optimal solutions of \eqref{main_opt}, each having a different number of pieces. In Figure \ref{fig:result}(d), we plot the optimal solution of the jointly constrained problem  \eqref{mod_opt}\footnote{We were unable to able find any additional optimal solutions for \eqref{mod_opt} for this example.}. The optimal solution is same as Figure \ref{fig:result}(c). It can be verified (numerically) that the solutions plotted in Figure \ref{fig:result}(a) and (b) are \emph{not} optimal solutions of \eqref{mod_opt}, highlighting the explicit difference in these formulations (and solutions they return) depending on weight and bias constraints. 

For comparison, in Figure \ref{fig:result}(d)-(g) we plot the output of single layer neural networks with different numbers $h$ of hidden neurons, obtained by more ``conventional'' training methods. Specifically, in each case the neural network weights and biases were initialized randomly with zero mean Gaussian random variables having standard deviation $10^{-3}$. The network was then trained, using a squared error loss, via gradient descent with step size (learning rate) $0.01$, until the loss was less than $10^{-4}$. As can be seen in the results, the output seems to approach the minimum norm solutions plotted in Figure \ref{fig:result}(c),(d), as the number of hidden neurons increases.

In Figure \ref{fig:result_bc}, we explore the extension to binary classification. The input points with same marker belong to the same class, and the boundary of the classifier is drawn in a solid (blue) line. Figure \ref{fig:result_bc}(a) shows the optimal classifier of \eqref{bc_opt} identified by CVX, and the jointly constrained version is shown in Figure \ref{fig:result_bc}(b). The remaining panels, Figures \ref{fig:result_bc}(c) and (d), show the output of neural networks with $h$ hidden neurons trained with same specifications as in the previous experiment except using logistic loss instead of squared error loss. As above, the solution of the trained examples appears to converge to that of the convex formulation as $h$ increases.

\section{Conclusions and Future Directions}
It is interesting to note that the atomic norm formulation employed here gave rise to optimization problems that share characteristics of \emph{group sparse} problems (see, e.g., \cite{yuan2006model, huang_gs}); here, the group structure manifests neuron-wise, and is determined by different sign patterns present in ${\mathcal J}$.  Whether insights from the group sparse literature can be employed here to provide additional insights (e.g., into efficient algorithmic methods, generalization results, etc.) is a topic for future work. 

Further, we note that the key insight in deriving the results presented here was to use the piecewise convex structure of the atomic set. This idea could presumably be extended to address multi-layer networks with piecewise activations, albeit at a cost of counting/bookkeeping.  We defer this investigation also to a future work.

\newpage
	\bibliographystyle{IEEEtran}
 	\bibliography{references}

\end{document}